%% file: mm.tex
\AtBeginDocument{%
  \providecommand\BibTeX{{%
    \normalfont B\kern-0.5em{\scshape i\kern-0.25em b}\kern-0.8em\TeX}}}

\documentclass[sigconf]{acmart}

\copyrightyear{2025}
\acmYear{2025}
\setcopyright{acmlicensed}
\acmConference[MM '25] {Proceedings of the 33rd ACM International Conference on Multimedia}{October 27--31, 2025}{Dublin, Ireland.}
\acmBooktitle{Proceedings of the 33rd ACM International Conference on Multimedia (MM '25), October 27--31, 2025, Dublin, Ireland}
\acmISBN{979-8-4007-2035-2/2025/10}
\acmDOI{XXXXXXX.XXXXXXX}

\usepackage{enumitem}
\usepackage{placeins}
\usepackage{cuted}
\usepackage{tikz}
\usepackage{comment}
\usepackage{enumitem}
\usepackage{color}
\usepackage{epsfig}
\usepackage{graphics}
\usepackage{color}
\usepackage{mathtools}
\usepackage{multirow}
\usepackage{tabularx}
\usepackage{amsfonts}
\usepackage{wrapfig,lipsum,booktabs}
\usepackage{xcolor}
\usepackage{floatflt}
\usepackage{subfig}
\usepackage{caption}
\usepackage{subfloat}
\usepackage{algpseudocode}
\usepackage{hyperref}
\usepackage{hyperxmp}
\usepackage[ruled, lined, linesnumbered, commentsnumbered, longend]{algorithm2e}

\usepackage{caption} 

\begin{document}

\setlength{\textfloatsep}{3pt}       
\setlength{\dbltextfloatsep}{3pt}    
\setlength{\intextsep}{3pt}          

\setlength{\floatsep}{3pt}           
\setlength{\dblfloatsep}{3pt}        

\setlength{\abovedisplayskip}{3pt}
\setlength{\belowdisplayskip}{3pt}
\setlength{\abovedisplayshortskip}{3pt}
\setlength{\belowdisplayshortskip}{3pt}
\title{ Segmenting Objectiveness and Task-awareness Unknown Region for Autonomous Driving}

\author{Mi Zheng}
\affiliation{%
  \institution{Harbin Institute of Technology}
  \city{Harbin}
  \country{China}}
\email{zhengrenjuzi@gmail.com}

\author{Guanglei Yang}
\authornotemark[1]
\affiliation{%
  \institution{Harbin Institute of Technology}
  \city{Harbin}
  \country{China}}
\email{
yangguanglei@hit.edu.cn}

\author{Zitong Huang}
\affiliation{%
  \institution{Harbin Institute of Technology}
  \city{Harbin}
  \country{China}}
\email{zitonghuang99@gmail.com}

\author{Zhenhua Guo}
\affiliation{%
  \institution{Tianyijiaotong Technology Ltd.}
  \city{Suzhou}
  \country{China}}
\email{zhenhua.guo@tyjt-ai.com}

\author{Kevin Han}
\affiliation{%
  \institution{Independent Researcher}
  \city{Menlo Park}
  \country{United States}}
\email{kevinwh@alumni.stanford.edu}

\author{Wangmeng Zuo}

\affiliation{%
  \institution{Harbin Institute of Technology}
  \city{Harbin}
  \country{China}}
\email{wmzuo@hit.edu.cn}



\input{0abstract}
\begin{strip}
\centering
\vspace{-3em} 
\includegraphics[width=\textwidth]{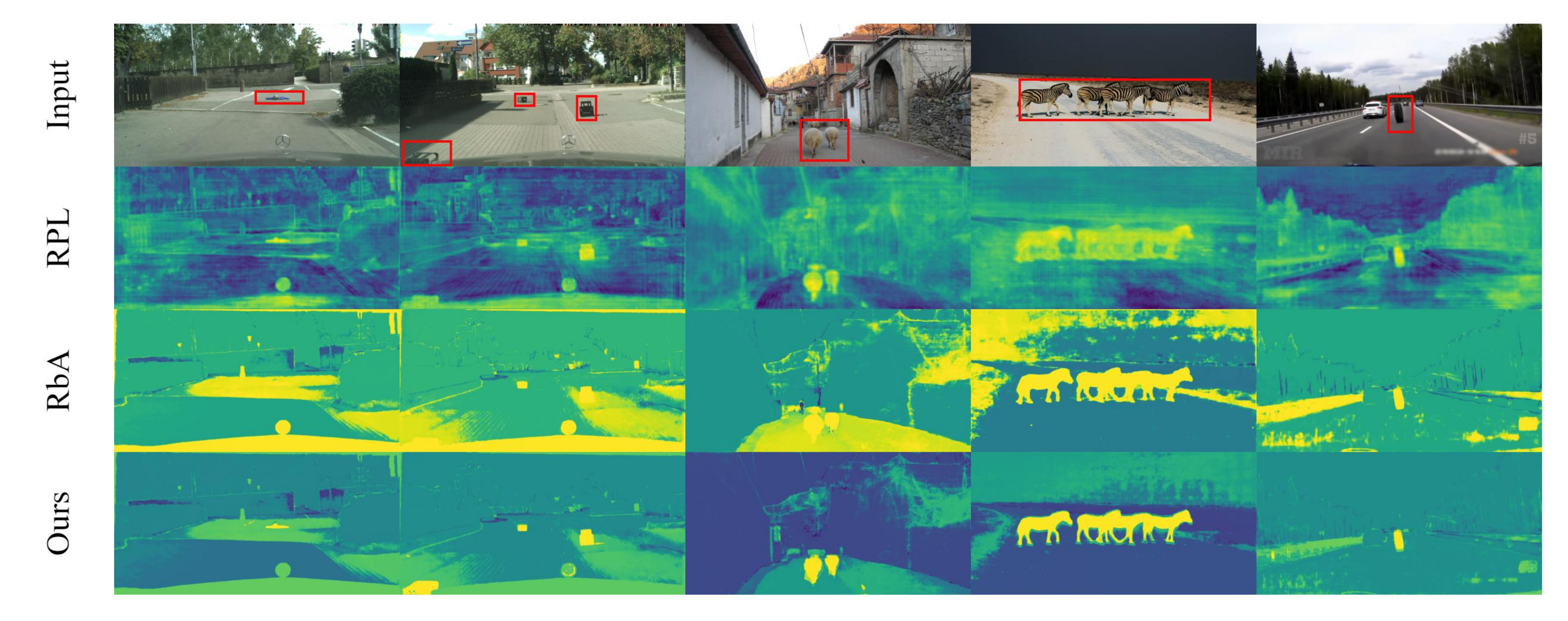}
\vspace{-3em}
\captionof{figure}{Visualisations for pixel-wise anomaly score maps. 
The top row shows real-world images
where the anomalous objects are highlighted by red boxes.
Subsequent rows show the anomaly score map for different methods, namely RPL~\cite{RPL}, RbA~\cite{nayal2023ICCV} and our approach SOTA.
Unlike other methods that often fail to fully recognize OOD objects and frequently introduce noise beyond the drivable area, SOTA accurately segments OOD objects while reducing noise outside the drivable area.}
\label{fig:large_figure}
\vspace{-1em}
\end{strip}
\maketitle

\input{1introduction}
\input{2relatedwork}

\input{3method}

\input{4experiments}

\input{5conclusions}

\section{Acknowledgements}
This work was supported by the National Key Research and Development
Program of China under Grant No. 2023YFA1008500.
Guanglei Yang is supported by the Postdoctoral Fellowship Program of CPSF under Grant Number GZC20233458 and the Postdoctoral Fund of Harbin Institute of Technology under Grant Number AUGA5630115523. 
\balance
{
\bibliographystyle{ACM-Reference-Format}
\bibliography{egbib}
}

\clearpage
\appendix
\section*{Appendix}   

\input{SM}

\end{document}

%% file: 0abstract.tex
\begin{abstract}

With the emergence of transformer-based architectures and large language models (LLMs), the accuracy of road scene perception has substantially advanced. 
Nonetheless, current road scene segmentation approaches are predominantly trained on closed-set data, resulting in insufficient detection capabilities for out-of-distribution (OOD) objects. 
To overcome this limitation, road anomaly detection methods have been proposed.
However, existing methods primarily depend on image inpainting and OOD distribution detection techniques, facing two critical issues: (1) inadequate consideration of the objectiveness attributes of anomalous regions, causing incomplete segmentation when anomalous objects share similarities with known classes, and (2) insufficient attention to environmental constraints, leading to the detection of anomalies irrelevant to autonomous driving tasks.
In this paper, we propose a novel framework termed Segmenting Objectiveness and Task-Awareness (SOTA) for autonomous driving scenes. 
Specifically, SOTA enhances the segmentation of objectiveness through a Semantic Fusion Block (SFB) and filters anomalies irrelevant to road navigation tasks using a Scene-understanding Guided Prompt-Context Adaptor (SG-PCA).
Extensive empirical evaluations on multiple benchmark datasets, including Fishyscapes Lost and Found, Segment-Me-If-You-Can, and RoadAnomaly, demonstrate that the proposed SOTA consistently improves OOD detection performance across diverse detectors, achieving robust and accurate segmentation outcomes.

\end{abstract}
\vspace{-2em}
\ccsdesc[500]{Computing methodologies~Scene anomaly detection}
\keywords{Anomaly Object Segmentation, Feature Fusion, Parameter-Efficient Fine-Tuning}

%% file: 1introduction.tex
\section{Introduction}
\label{sec:intro}
The integration of transformer-based architectures ~\cite{setr2021,maskformer2021,mask2former2022,dpt2021,segmenter2021}and large-scale pretrained models~\cite{swin2021,sam2023,dosovitskiy2020image} has significantly advanced semantic segmentation in road scene perception. These methods have shown outstanding performance in closed-set environments, characterized by predefined object categories like vehicles and pedestrians~\cite{cityscapes}. However, real-world environments are inherently open, often containing unexpected out-of-distribution (OOD) objects—from accident debris to rare natural phenomena—that significantly challenge traditional segmentation models. Misclassifying these anomalies as background or assigning them to known categories introduces substantial safety hazards, particularly in autonomous driving. For instance, failing to detect a fallen tree or a misplaced construction sign can result in severe accidents. Consequently, reliably identifying and segmenting OOD objects without compromising in-distribution (ID) segmentation accuracy has emerged as a critical challenge, termed road anomaly detection, for robust and safe scene understanding~\cite{yang2021OODsurvey,bogdoll2022anomaly}.

Current road anomaly detection methods follow two main technical routes. The first paradigm relies primarily on reconstruction-based approaches~\cite{lis2023detecting,zavrtanik2020reconstruction,xia2020synthesize,lis2019detecting,wang2018highres}, which use generative models to learn normal scene distributions and detect anomalies via high reconstruction errors. These methods proved effective in controlled environments but faced inherent challenges in complex real-world scenarios. Recently, the field has shifted towards score-based methods~\cite{PEBAL,Synboost,RPL,nayal2023ICCV,rai2023unmasking,delic24bmvc}, framing anomaly detection as a confidence estimation problem. These approaches typically leverage prediction uncertainty or feature-space distances to compute pixel-level anomaly scores. Representative works include RPL~\cite{RPL}and RbA~\cite{rai2023unmasking}. RPL~\cite{RPL} uses residual pattern learning to identify deviations from expected feature distributions, while RbA~\cite{nayal2023ICCV} employs a multi-head rejection mechanism, treating mask classification results as multiple one-vs-all classifiers for robust anomaly identification.

Despite significant advancements, prior approaches often treat road anomaly detection as a general anomaly detection problem, where all outliers are deemed anomalous. This leads to two fundamental limitations hindering their practical deployment in autonomous driving systems, as shown in Fig.~\ref{fig:large_figure}. First, they struggle with incomplete segmentation of partial anomalies. Current methods fail to capture all OOD-specific pixels, especially when anomalies share partial visual characteristics with in-distribution objects. For example, state-of-the-art RbA~\cite{nayal2023ICCV} (Fig.~\ref{fig:large_figure}, column 1–2) detects anomalies like damaged tires but fails to segment them entirely, leaving residual regions misclassified as background. This issue arises because its multi-head rejection mechanism suppresses uncertain pixels without explicitly modeling OOD shape continuity, leading to fragmented masks with undetected anomaly parts. Second, they suffer from task-agnostic overdetection. Existing methods overlook road scene constraints, leading to false alarms in irrelevant regions. As shown in Fig.~\ref{fig:large_figure} (column 3–5), non-critical objects outside drivable areas, such as roadside vegetation and distant buildings, are erroneously flagged as OOD. This happens because RbA’s\cite{nayal2023ICCV} threshold-based scoring system prioritizes anomaly likelihood without considering autonomous driving task requirements, resulting in overdetection in areas unrelated to navigation safety. These two issues—partial under-segmentation and task-irrelevant overdetection—highlight the need for anomaly detection frameworks that explicitly model OOD objective attributes while incorporating scene constraints to enhance autonomous driving reliability.

In this paper, we propose Segmenting Objectiveness and Task-awareness (SOTA), a unified framework integrating semantic feature fusion with scene-understanding guided prompt learning. It comprises two modules: the Semantic Fusion Block (SFB) and the Scene-understanding Guided Prompt-Context Adapter (SG-PCA). For SFB, the pixel-wise segmentor's vanilla OOD prediction is aligned to the latent vision space via projection and alignment blocks. Fusing the OOD map with vision features improves objectiveness segmentation precision by refining OOD-specific attribute detection and reducing partial anomaly errors. SG-PCA, conversely, extracts road scene priors (e.g., lane topology, drivable areas) through task-aware aggregation, adapting Erosion and Dilation operations to resolve partial occlusion issues. Subsequently, scene priors and the vanilla OOD prediction combine to generate task-aware prompts via multi-aware cross-attention. By suppressing navigation-irrelevant anomalies through prompt learning, SG-PCA ensures focus on safety-critical regions, overcoming environmental constraints inherent in distribution-based methods. Moreover, we introduce parameter-efficient adaptation of SAM’s~\cite{sam2023} mask decoder using Low-Rank Adaptation (LoRA)~\cite{hu2022lora}, enabling seamless integration of enriched OOD embeddings without additional manual threshold tuning or postprocessing. Our contributions are summarized as follows:
\begin{itemize}
\item We are the first to incorporate both objectiveness and task-awareness into road anomaly detection, significantly improving its practical utility for real-world autonomous driving applications.
\item We propose Segmenting Objectiveness and Task-awareness (SOTA), comprising two key modules—Semantic Fusion Block (SFB) and Scene-understanding Guided Prompt-Context Adapter (SG-PCA)—that together achieve spatial and semantic completeness in anomaly detection.
\item Extensive experiments on benchmark datasets demonstrate that our approach significantly outperforms the baseline and state-of-the-art method in both pixel-level and component-level anomaly segmentation results, achieving marked improvements across key evaluation metrics.
\end{itemize}
\vspace{-10pt}

%% file: 2relatedwork.tex
\section{Related Work}

\noindent\textbf{Semantic Segmentation.}

Semantic segmentation, the task of assigning a class label to each pixel in an image, is critical in scene understanding and computer vision~\cite{cityscapes,thisanke2023semantic,minaee2020image}. Early models~\cite{long2015fully,ronneberger2015u,badrinarayanan2017segnet,zhao2017pyramid,chen2018encoder} leveraged convolutional neural networks (CNNs) for pixel-wise classification, extracting hierarchical features across scales. Building on this, methods like DeepLab~\cite{chen2017deeplab} used dilated convolutions for expanded receptive fields without sacrificing resolution, while DeepLabv3+~\cite{chen2018encoder} refined segmentation by combining atrous convolution with a decoder, improving boundary delineation and object accuracy.

More recently, transformer-based architectures~\cite{dosovitskiy2020image,swin2021} have shown the ability to model long-range dependencies and global context, significantly improving over CNNs. Frameworks like MaskFormer~\cite{maskformer2021} and Mask2Former~\cite{mask2former2022}, integrating mask-based attention, have unified various segmentation tasks and achieved state-of-the-art performance. Alongside these, prompt-based segmentation, exemplified by the Segment Anything Model (SAM)~\cite{sam2023}, enables more flexible and accurate segmentation via prompts like points, boxes, or text. These developments collectively advance the field, driving robust semantic segmentation in increasingly complex real-world scenarios.

However, traditional semantic segmentation models, trained on closed domains, rely on predefined categories. They are not designed for out-of-distribution (OOD) objects or unknown categories. Thus, they struggle in open-set environments like autonomous driving, where new or unexpected objects (e.g., debris, rare animals) may appear. For driving safety, semantic segmentation must detect such OOD objects, ensuring reliable scene understanding and robust performance in dynamic, open-world scenarios. This limitation has shifted research focus to road anomaly detection, a critical area addressing these challenges by enabling OOD object detection and segmentation in driving environments.

\noindent\textbf{Anomaly Segmentation in autonomous driving.}

Anomaly segmentation involves identifying image regions deviating from normal patterns, often for detecting out-of-distribution (OOD) objects or unusual events~\cite{bogdoll2022anomaly, yang2021OODsurvey}. This task extends traditional semantic segmentation by classifying pixels and recognizing anomalies, making it particularly useful in autonomous driving. Early methods relied on probabilistic models to compute uncertainty scores (e.g., Monte-Carlo dropout, deep ensembles)\cite{hendrycks2016baseline, lakshminarayanan2017simple, mukhoti2018evaluating, jung2021standardized} and reconstruction-based approaches\cite{lis2023detecting, zavrtanik2020reconstruction, xia2020synthesize, lis2019detecting, wang2018highres}, which detect anomalies by measuring reconstruction failure.

Recent advancements~\cite{delic24bmvc, grcic23cvprw, nayal2023ICCV,rai2023unmasking,RPL,liu2025ooddinoamultilevelframeworkanomaly} have shifted towards more sophisticated models integrating deep learning with uncertainty estimation. For example, both RbA~\cite{nayal2023ICCV} and Mask2Anomaly~\cite{rai2023unmasking} use mask-class pairs; RbA derives anomaly scores for regions not covered by masks, while Mask2Anomaly transitions from per-pixel to mask classification, employing global masked attention and mask contrastive learning. Additionally, RPL~\cite{RPL} detects OOD pixels while minimizing impact on inlier segmentation accuracy. Several works also adapt open-set segmentation methods to detect unknown anomalies, leveraging transformer-based architectures or prompt-based models like SAM~\cite{sam2023}. These methods, including S2M~\cite{zhao2024segment}, show significant performance gains; however, current approaches using SAM only apply it during inference and lack training or fine-tuning strategies to optimize SAM for anomaly detection.

However, existing anomaly segmentation methods typically focus on distribution-based considerations, overlooking the unique characteristics of the driving environment. In real-world applications, detected anomalies must correspond to meaningful objects or structures, not arbitrary regions. Furthermore, their relevance to the driving task, particularly safety impact, must be considered. In this paper, we propose Segmenting Objectiveness and Task-awareness (SOTA), a method designed to segment anomaly objects with both spatial and semantic awareness. This approach ensures anomalies are identified based on objectivity and evaluated for driving task relevance, thereby improving segmentation accuracy and the system’s ability to prioritize safety-critical regions.

%% file: 3method.tex
\begin{figure*}[!t]
\begin{center}
\includegraphics[width=\linewidth]{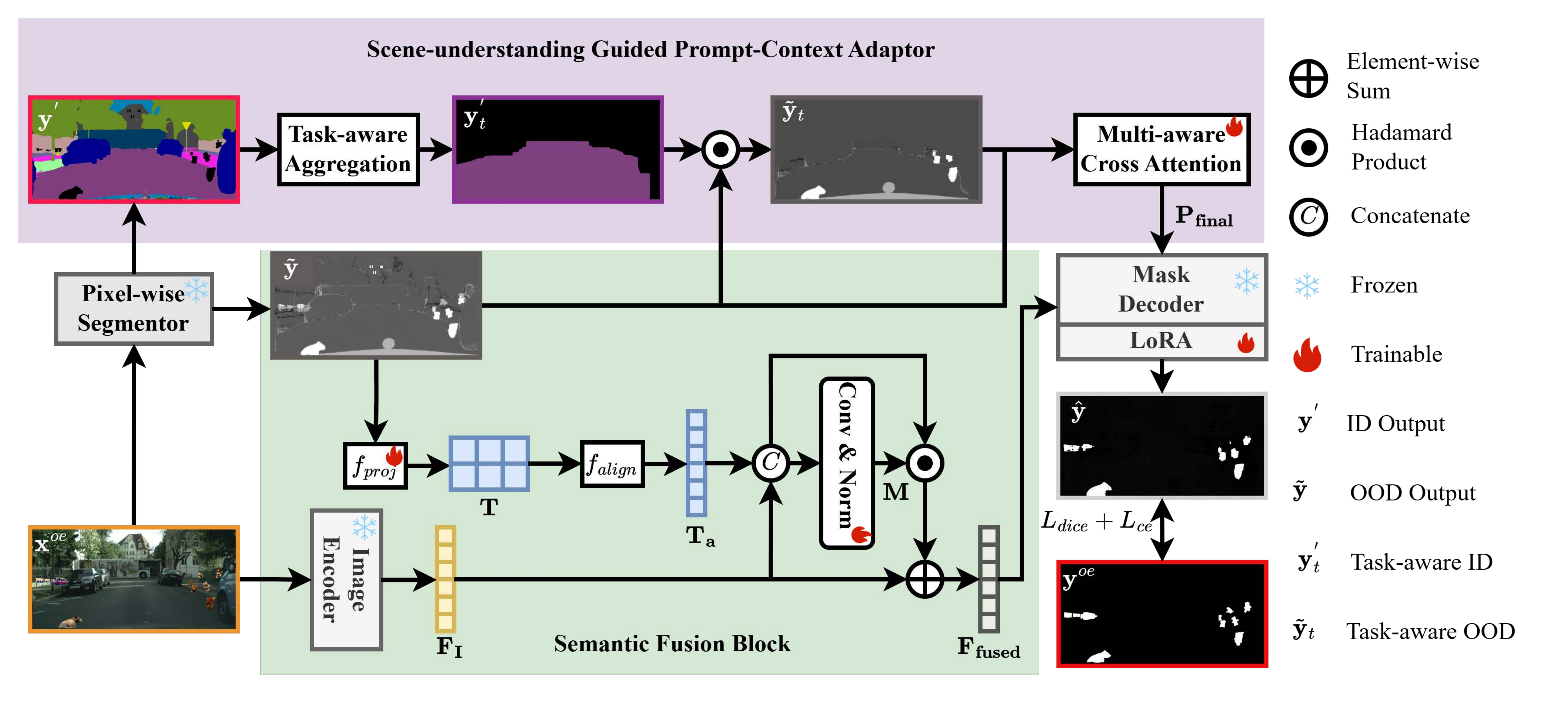}
\end{center}
\vspace{-2em}
  \caption{\textbf{Overview of our SOTA.} Our approach comprises two principal modules: (a). The Semantic Fusion Block (SFB), highlighted by the green regions, fuses image features with pixel-level anomaly scores  \(\tilde{\mathbf{y}}\) to emphasize out-of-distribution cues; (b). The Scene-understanding Guided Prompt Learning (SG-PCA) module is highlighted by the purple regions in our design. First, it applies task-aware aggregation to extract context-specific prompt, denoted as \(\mathbf{y'}_t\). Then, hadamard product and multi-aware cross attention is used to fuse \(\mathbf{y'}_t\) with the raw anomaly scores \(\tilde{\mathbf{y}}\), integrating detailed scene context.
 The fused feature and refined prompt are then input to a LoRA-adapted mask decoder, yielding precise anomaly segmentation \(\hat{\mathbf{y}}\).}
\label{fig:overview}
\end{figure*}

\vspace{-0.7ex}
\section{Methodology}
\label{sec:method}

Our approach builds upon a pre-trained pixel-wise OOD detector that generates  OOD confidence maps. First, we introduce the Semantic Fusion Block (SFB) , which combines image embeddings from the Segment Anything Model (SAM)~\cite{sam2023} with  OOD confidence maps. This fusion ensures that anomalies are effectively highlighted without overshadowing the core in-distribution features. Next, to capture road scene priors, we design the Scene-understanding Guided Prompt-Context Adapter (SG-PCA), which utilizes multi-aware cross-attention to extract and integrate relevant contextual information. Finally, for seamless integration of enriched OOD embeddings without postprocessing, we adopt Low-Rank Adaptation (LoRA)~\cite{hu2022lora} in the mask decoder, ensuring a parameter-efficient incorporation of these enriched embeddings.


\vspace{-1ex}
\subsection{Preliminaries}
\label{subsec:preliminaries}

Let \(\mathbf{x^{oe}} \subset \mathbb{R}^{H \times W \times 3}\) denote the space of input RGB images, where \(H\) and \(W\) represent the image height and width. The goal of anomaly segmentation is to obtain a prediction map \(\hat{\mathbf{y}} \subset \mathbb{R}^{H \times W}\), in which higher values indicate a higher probability that a pixel belongs to an out-of-distribution (OOD) region.
Most existing anomaly segmentation methods~\cite{nayal2023ICCV,rai2023unmasking,delic24bmvc} are built upon a semantic segmentation network. Let \(\mathbf{y'} \in \mathbb{R}^{C \times H \times W}\) denote the output of such a network, where \(C\) is the number of in-distribution (ID) classes. For example in RbA~\cite{nayal2023ICCV},the OOD score is first computed directly from \(\mathbf{y'}\) by applying a normalization function \(\sigma\) (e.g., \(\tanh\) or softmax) to obtain per-class confidence. Then they aggregate the inlier confidence across all \(C\) classes as\(
\sum_{k=1}^{C} \sigma\bigl(\mathbf{y'_k(x)}\bigr)
\), where each element \(\mathbf{y'_k(x)}\) represents the raw logit score for the \(k\)th class at pixel \(x\), indicating the unnormalized confidence that pixel \(x\) belongs to class \(k\). 
The OOD score is  defined as the complement of this sum:
\begin{align}
\mathbf{\tilde{y}(x)} = 1 - \sum_{k=1}^{C} \sigma\bigl(\mathbf{y'_k(x)}\bigr),
\end{align}
where a higher value of \(\mathbf{\tilde{y}(x)}\) indicates a lower confidence in any known class and thus a higher likelihood of being an outlier region.
This formulation interprets low-confidence regions across all inlier classes as potential anomalies. While effective for isolated anomalies, this paradigm exhibits critical limitations when handling the whole OOD objects and contextual road semantics, as analyzed in Section~\ref{sec:intro}.

Our proposed method SOTA addresses these limitations by integrating semantic fusion and scene-aware prompt learning. As illustrated in Figure\ref{fig:overview}, we leverage SAM’s robust segmentation capabilities to ensure complete object segmentation. In addition, our Semantic Fusion Block (SFB) augments SAM’s ability to segment OOD objects by fusing the OOD semantic information from the pixel-wise segmentor (e.g., RbA) with the semantic features extracted by SAM’s image encoder. Concurrently, the Scene-understanding Guided Prompt-Context Adapter (SG-PCA) extracts road scene priors using task-aware aggregation to generate robust scene prompts. Anomaly scores and scene priors are then integrated via multi-aware cross-attention to produce task-aware prompts that effectively suppress irrelevant anomalies. Finally, we employ a parameter-efficient fine-tuning strategy using Low-Rank Adaptation (LoRA) to adapt SAM's mask decoder, ensuring seamless integration of the enhanced image embeddings and task-aware prompt embeddings. This unified framework not only overcomes the challenge of incomplete anomaly segmentation but also leverages scene context for improved OOD detection, as discussed in the following sections.

\vspace{-1em}
\subsection{Semantic Fusion Block}
\label{subsec:SFBlock}

Our Semantic Fusion Block is designed to seamlessly combine anomaly cues derived from tokenized OOD features with the primary image embeddings, thereby enhancing the contextual accuracy of anomaly segmentation. Given an outlier-exposed image \(\mathbf{x^{oe}} \in \mathbb{R}^{H \times W \times 3}\), the SAM's image encoder extracts the main image feature map 
\(\
\mathbf{F_{I}} \in \mathbb{R}^{D_I \times H' \times W'},
\)
where \(D_I\) is the image feature dimension and \(H'\times W'\) is the spatial resolution. In parallel, a pixel-level segmentor (e.g., RbA) produces an OOD confidence map 
\(\mathbf{\tilde{y}} \in [0,1]^{H \times W}\). To transform this single-channel anomaly confidence map into a rich latent representation, we first project \(\tilde{y}\) using a learnable projection function \(f_{\mathrm{proj}}(\cdot)\), yielding a token feature 
\(
\mathbf{T} \in \mathbb{R}^{D_T \times H' \times W'},
\)
with \(D_T\) being the token dimension. To ensure compatibility between the token representation and the image features, we further align \(\mathbf{T}\) with \(\mathbf{F_I}\) via an alignment function \(f_{\mathrm{align}}(\cdot)\), resulting in an aligned token
\begin{equation}
\mathbf{T_a} = f_{\mathrm{align}}(\mathbf{T}). 
\end{equation}
Subsequently, we concatenate the aligned token \(\mathbf{T_a}\in \mathbb{R}^{D_I \times H' \times W'}\) with the image embedding \(\mathbf{F_I}\) along the channel dimension to obtain a composite feature tensor:
\begin{equation}
\mathbf{F_c} = \mathrm{Concat}(\mathbf{T_a}, \mathbf{F_I}).
\end{equation}
To dynamically integrate these heterogeneous features, we apply an attention mechanism. In detail, we first generate an attention weight map \(\mathbf{M} \in \mathbb{R}^{H' \times W'}\) using a lightweight convolutional layer followed by a sigmoid activation:
\begin{equation}
    \mathbf{M} = \sigma\bigl(\mathrm{Conv}(\mathbf{F_c})\bigr).
\end{equation}

Then, the final fused embedding is computed by reweighting the composite features with \(\mathbf{M}\) and incorporating a residual connection from the original image embedding:
\begin{equation}
\mathbf{F_{\mathrm{fused}}} = (1 + \mathbf{M}) \odot \mathbf{F_I} + (1 - \mathbf{M}) \odot \mathbf{T_a} . 
\end{equation}
This fusion process ensures that the anomaly cues are selectively and adaptively integrated with the image features. The resulting embedding 
\(\mathbf{F_{\mathrm{fused}}}\) highlights regions with high OOD likelihood while preserving the dominant semantic context, and is fed to the mask decoder for accurate anomaly segmentation.

\subsection{ Scene-understanding Guided Prompt Context Adapter}
\label{subsec:SG-PCA}

To enhance both the precision and the contextual relevance of anomaly segmentation, we propose a  process that refines the anomaly prompt by leveraging scene understanding and a context adapter. In this framework, the process is divided into two complementary components: (1) a Task-aware Aggregation module that exploits scene cues (specifically the road region) to generate a semantically meaningful prompt, and (2) a Multi-aware Cross Attention module that employs road-wise cross attention to dynamically integrate contextual features from the anomaly cues.

The first step is dedicated to harnessing the scene context, especially the drivable road area, to constrain where anomalies are likely to appear. Our task-aware aggregation module extracts a preliminary road mask \( \mathbf{y}_t \in \{0,1\}^{H \times W}\) from the semantic segmentation output \(\mathbf{y}' \in \mathbb{R}^{C \times H \times W}\) by assigning 1 to pixels classified as road and 0 otherwise. Due to misclassification and the presence of other objects, the extracted road mask \( \mathbf{y}_t \) often contains spurious holes and discontinuities. To address this, we refine \( \mathbf{y}_t \) through morphological operations. Specifically, we apply an erosion followed by a dilation~\cite{beucher1991watershed}, which can be expressed as
\begin{align}
\mathbf{y}'_t = \mathrm{Dilate}\bigl(\mathrm{Erode}(\mathbf{y}_t, K, n), K, n\bigr),
\end{align}
where \(K\) denotes a \(k \times k\) structuring element and \(n\) is the number of iterations, ensuring that small holes are filled and the road region is more completely delineated.

Once the refined road prompt is obtained, it is integrated with the raw out-of-distribution (OOD) confidence map \( \tilde{\mathbf{y}} \). This integration is performed via a Hadamard product:
\begin{equation}
\tilde{\mathbf{y}}_{t} = \tilde{\mathbf{y}} \odot \mathbf{y}'_t,
\label{eq:refined_prompt}
\end{equation}
where the multiplication serves to restrict the anomaly scores to regions that are consistent with the scene context (i.e., predominantly within or near the drivable region). This step not only reduces false positives from areas outside the road but also reinforces the importance of the drivable area as a reference for potential anomalies.

In practice, even a well-refined road prompt might not capture the full complexity of out-of-distribution objects, particularly when such objects extend partially into non-drivable regions due to perspective or occlusion. To tackle this limitation, our second stage introduces a road-wise cross attention that employs cross attention to more fully integrate the raw anomaly signals with the scene-informed prompt.

Specifically, we begin by projecting both the task-guided anomaly map \( \tilde{\mathbf{y}}_{t} \) and the original anomaly confidence map \( \tilde{\mathbf{y}} \) into a shared embedding space. This projection is achieved via learnable 1×1 convolutional layers denoted as \( f_Q \), \( f_K \), and \( f_V \), which produce the query, key, and value representations, respectively:
\begin{equation}
\begin{aligned}
\mathbf{Q} &= f_Q(\tilde{\mathbf{y}}_{t}), \quad \mathbf{K} = f_K(\tilde{\mathbf{y}}), \quad \mathbf{V} = f_V(\tilde{\mathbf{y}}),
\end{aligned}
\label{eq:QKV}
\end{equation}with \(\mathbf{Q}, \mathbf{K}, \mathbf{V} \in \mathbb{R}^{d \times H' \times W'}\) and \(d\) being the common embedding dimension (and \(H'\times W'\) the spatial resolution after appropriate resizing). 

The road-wise cross attention mechanism then computes an attention matrix \(\mathbf{A} \in \mathbb{R}^{d \times d}\) by aligning the task-guided features with the raw anomaly features:
\begin{equation}
\mathbf{A} = \mathrm{Softmax}\!\left(\frac{\mathbf{QK^T}}{\sqrt{d}}\right),
\label{eq:attention}
\end{equation}
with the softmax function applied over the channel dimension. This attention matrix \( \mathbf{A} \) captures the inter correlations, reflecting how road area of the refined prompt relate to the original anomaly map.

Finally, the attention matrix is used to reweight the value representation, yielding the final prompt embedding:
\begin{equation}
\mathbf{P}_{\mathrm{final}} = \mathbf{A} \cdot \mathbf{V},
\label{eq:final_prompt}
\end{equation}
The resulting embedding fuses the spatial constraints from the refined road prompt with comprehensive anomaly information, thereby enabling the mask decoder to achieve robust and precise segmentation of out-of-distribution objects. This anomaly detection in scene context through road extraction and enhancing it with road-wise cross attention ensures robustness to variations such as perspective-induced partial object appearances while maintaining high segmentation accuracy.

\input{tab/ra_fs_laf}

\subsection{Training and Inference Pipeline}
\label{subsec:LoRA}
In our framework, the mask decoder in the Segment Anything Model (SAM)~\cite{sam2023} receives both the fused image features \(\mathbf{F_{\mathrm{fused}}}\) from our Semantic Fusion Block and the context-aware prompt embedding \(\mathbf{P}_{\mathrm{final}}\) from the Scene-understanding Guided Prompt Context Adapter, and produces the final segmentation mask \(\mathbf{\hat{y}} \). The final segmentation mask is computed as
\begin{align}
   \mathbf{\hat{y}} = \mathrm{MaskDecoder}\Bigl(\mathbf{F_{\mathrm{fused}}}, \mathbf{P}_{\mathrm{final}};\theta\Bigl), 
\end{align}
where \(\theta\) denotes the mask decoder's parameters. The final result integrates both the enriched image features and the context-aware prompt. Because the image embedding now incorporates out-of-distribution (OOD) confidence information along with the original features, and because the prompt embedding further provides rich contextual cues from the scene, the mask decoder  must be adapted to fully exploit these enriched representations. To achieve efficient adaptation of the pre-trained mask decoder while preserving overall performance, we adopt Low-Rank Adaptation (LoRA)~\cite{hu2022lora}. This parameter-efficient adaptation allows the mask decoder to effectively leverage the additional anomaly cues without requiring extensive retraining or manual threshold tuning.

Although LoRA specifically targets the decoder’s parameters, the entire pipeline is trained end-to-end so that earlier components can also adjust to the new distribution of features. Let  $\mathbf{y}^{oe}$ be the ground-truth mask for OOD objects. We employ a combined Dice and cross-entropy loss~\cite{milletari2016vnet} to supervise the entire network:
\begin{align}
\mathcal{L}
\;=\;
\,\mathcal{L}_{\mathrm{Dice}}(\hat{\mathbf{y}},\,\mathbf{y}^{oe})
\;+\;
\,\mathcal{L}_{\mathrm{CE}}(\hat{\mathbf{y}},\,\mathbf{y}^{oe}).
\end{align}
By optimizing this objective, the decoder adapts to the enriched embedding while the preceding modules refine their outputs to better highlight anomalies, leading to more accurate segmentation of unknown objects in open-set scenarios.

During inference, the input image is simultaneously processed by SAM's image encoder and a pixel-level segmentor. The encoder produces image features, while the segmentor generates an out-of-distribution (OOD) confidence map and an initial semantic segmentation for known categories. Our framework then performs three key operations: first, the image features are fused with the OOD confidence map to obtain a refined fused embedding that emphasizes anomalous signals without losing the core inlier representation; second, the semantic segmentation and OOD confidence map are fed into our scene-understanding guided prompt learning module to generate a context-aware prompt; finally, the finetuned mask decoder receives both the fused embedding and the context-aware prompt, and its output is merged with the original OOD confidence map to yield the final anomaly segmentation result.

%% file: tab/ra_fs_laf.tex
\begin{table*}[t!]
    \footnotesize
    \centering
    \caption{\textbf{Pixel-level results.} 
    We show the results with the best and second-best results in \textbf{bold} and \underline{underlined}.Our method is based on the anomaly score from RbA.
    SOTA notably improves the results in all metrics across datasets.}
    \vspace{-1.7em}
    \resizebox{\textwidth}{!}{%
    \begin{tabular}{r c c c c c c c c c c} 
        \toprule
        \multicolumn{1}{c}{\multirow{2}{*}{Method}} & \multicolumn{3}{c}{Road Anomaly} & \multicolumn{3}{c}{Fishyscapes L\&F} & \multicolumn{2}{c}{SMIYC RA-21} & \multicolumn{2}{c}{SMIYC RO-21} \\
        \cmidrule(r){2-4} \cmidrule(l){5-7} \cmidrule(l){8-9} \cmidrule(l){10-11}
         & AuROC $\uparrow$ & AuPRC $\uparrow$ & FPR $\downarrow$ & AuROC $\uparrow$ & AuPRC $\uparrow$ & FPR $\downarrow$ & AuPRC $\uparrow$ & FPR $\downarrow$ & AuPRC $\uparrow$ & FPR $\downarrow$ \\
        \midrule
        Max Softmax~\cite{hendrycks2016baseline}  & 73.76 & 20.59 & 68.44 & 86.99 &  6.02 & 45.63 & 40.40 & 60.20 & 43.40 & 3.80 \\
        Entropy~\cite{hendrycks2016baseline}       & 75.12 & 22.38 & 68.15 & 88.32 & 13.91 & 44.85 & - & - & - & - \\
        Mahalanobis~\cite{lee2018simple}           & 76.73 & 22.85 & 59.20 & 92.51 & 27.83 & 30.17 & 22.50 & 86.40 & 25.90 & 26.10 \\
        SML~\cite{jung2021standardized}            & 81.96 & 25.82 & 49.74 & 96.88 & 36.55 & 14.53 & 21.68 & 84.13 & 18.60 & 91.31 \\ 
        Maximized Entropy~\cite{chan2021entropy}     &   -   &   -   &   -   & 93.06 & 41.31 & 37.69 & 80.70 & 17.40 & 94.40 & 0.40 \\
        SynBoost~\cite{Synboost}                   & 81.91 & 38.21 & 64.75 & 96.21 & 60.58 & 31.02 & 68.8 & 30.9 & 81.4 & 2.8 \\
        GMMSeg~\cite{liang2022gmmseg}              & 89.37 & 57.65 & 44.34 & 97.83 & 50.03 & 12.55 & - & - & - & - \\
        PEBAL~\cite{PEBAL}                         & 88.85 & 44.41 & 37.98 & 98.52 & 64.43 &  6.56 & 53.10 & 36.74  & 10.45 & 7.92 \\ 
        RPL+CoroCL~\cite{RPL}                      & 95.72 & 71.60 & 17.74 & \textbf{99.39} & 70.61 & \textbf{2.52} & 88.55 & 7.18 & 96.91 & 0.09 \\
        Mask2Anomaly~\cite{rai2023unmasking}        & 96.57 & 79.53 & 13.54 & 95.40 & 69.43 &  9.18 & \underline{94.50} & \underline{3.30} & 88.60 & 0.30 \\
        \midrule
        RbA~\cite{nayal2023ICCV}                  & \underline{97.99} & \underline{85.42} & \underline{6.92} & 98.62 & \underline{70.81} & 6.30 & 91.54 & 5.82 & \underline{98.24} & \underline{0.041} \\
        \textbf{SOTA (Ours)}                        & \textbf{98.71} & \textbf{92.46} & \textbf{4.03} & \underline{98.92} & \textbf{76.10} & \underline{3.53} & \textbf{96.39} & \textbf{1.90} & \textbf{98.26} & \textbf{0.04} \\
        \bottomrule
    \end{tabular}}
    \label{tab:ra_fs_val}
\end{table*}

%% file: 4experiments.tex
\section{Experiments}


\subsection{Experimental Setup}
\noindent\textbf{Datasets.}  
We evaluate our approach on several widely used OOD detection benchmarks. Specifically, we employ the Segment-Me-If-You-Can (SMIYC) benchmark~\cite{segmentmeifyoucan}, which includes two datasets: RoadAnomaly (RA) with 110 images exhibiting diverse anomalies under challenging conditions, and RoadObstacle (RO) with 442 images focusing on small obstacles. Additionally, experiments are conducted on the Road Anomaly dataset~\cite{lis2019detecting}—comprising 60 real-world images—and on the Fishyscapes Lost\&Found dataset~\cite{blum2021fishyscapes}, which provides 100 validation and 275 test images that emphasize small, less diverse anomalies.

\noindent\textbf{Outlier Exposure.}
Following prior works~\cite{nayal2023ICCV, RPL, PEBAL}, we adopt an OE strategy to enrich the training distribution. Cityscapes~\cite{cityscapes} images are used only as unlabeled backgrounds, without exploiting their semantic annotations. Object instances from COCO~\cite{COCO}, extracted with their ground-truth masks, are composited into Cityscapes scenes to simulate anomalous appearances. Although COCO supervision is used to obtain object regions, no inlier supervision from Cityscapes is involved; hence, the conventional IND–OOD supervision paradigm is not explicitly applied. Instead, COCO simply provides additional object diversity, enabling the model to generalize better to unexpected visual patterns in real-world driving scenarios.
\input{tab/component}

\noindent\textbf{Implementation Details.}  
Our approach builds upon the framework of RbA~\cite{nayal2023ICCV}, leveraging its anomaly score and inlier logits for both training and inference. The segmentation backbone utilized in RbA is Mask2Former~\cite{mask2former2022}, equipped with a Swin-B~\cite{swin2021} transformer encoder. To further enhance segmentation quality, we integrate the SAM as a promptable segmentation component, employing its ViT-H backbone for improved feature representation.
During training, we initialize the learning rate at $1\times 10^{-4}$ and employ a polynomial learning rate decay strategy, where the learning rate follows the schedule $1-\frac{\text{iter}}{\text{max\_iter}}$. For further details, please refer to the Supplementary Materials.
\begin{figure*}[t]
  \centering
  \includegraphics[ height=0.2\textheight, keepaspectratio]
{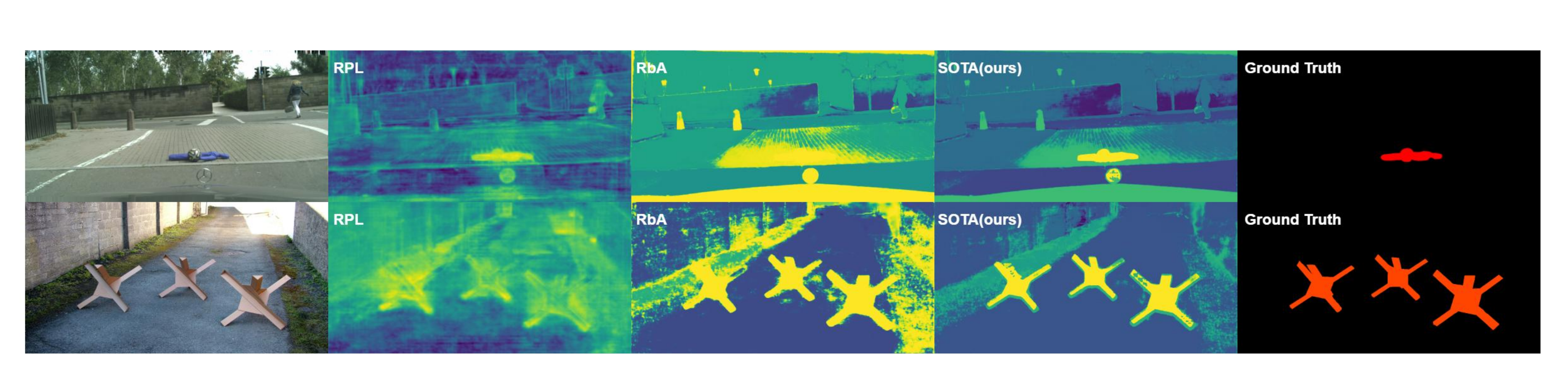} 
  \vspace{-3em}
  \caption{Qualitative Results. Visualization comparisons of RPL~\cite{RPL}, RbA~\cite{nayal2023ICCV}, and our proposed method. Our approach demonstrates superior segmentation quality by providing more complete and accurate identification of OOD objects while effectively reducing redundant pixel predictions beyond the road region.}
  \label{fig:qualitaty}
\end{figure*}

\noindent\textbf{Evaluation Metrics.}  
To facilitate a comprehensive comparison with previous methods, we report standard evaluation metrics used in OOD segmentation benchmarks. For pixel-wise evaluation, we evaluate model performance using Area under the Precision-Recall Curve (AuPRC) , the Area Under the Receiver Operating Characteristic Curve (AuROC), and the False Positive Rate at 95\% True Positive Rate (FPR@95). For component-level evaluation, we follow the SMIYC benchmark's official evaluation protocol. Specifically, the SMIYC benchmark reports the averaged component-wise F1 score (\(F1^*\)), the positive predictive value (PPV), and the segmentation intersection-over-union (sIoU). These metrics quantify the number of true positives (TP), false negatives (FN), and false positives (FP) at the instance level, providing a holistic evaluation of anomaly segmentation quality.

\subsection{Main Results}

\noindent\textbf{Pixel-level Results.}  
As summarized in Table~\ref{tab:ra_fs_val}, our method significantly outperforms the baseline RbA~\cite{nayal2023ICCV} and achieves state-of-the-art performance across multiple datasets. While the RPL+CoroCL\cite{RPL} achieves slightly better AuROC and FPR on the Fishyscapes Lost\&Found dataset, SOTA improves the AuPRC by a significant margin of 5.5\% compared to RPL+CoroCL. Furthermore, SOTA consistently outperforms RPL+CoroCL across all other datasets.

\input{tab/module_ablation}



\noindent\textbf{Component-level Results.}  
Table~\ref{tab:comp-eval} presents the component-level evaluation on the SMIYC RA-21 and SMIYC RO-21 benchmarks. Compared to RbA, our method achieves substantial improvements on RA-21, including a 5.58\% increase in sIoU, an 18.31\% boost in PPV, and a 14.67\% gain in F1*. Similarly, on RO-21, our method yields an 8.00\% improvement in PPV and a 9.55\% increase in F1*. While Mask2Anomaly achieves a higher performance on RO-21, our approach surpasses it on RA-21, achieving a 0.88\% improvement in sIoU and a 12.87\% increase in F1*. Moreover, while ContMAV achieves higher PPV and F1* scores on RA-21, it does not report results on RO-21, whereas our approach consistently enhances performance across both datasets.

\begin{figure}[h!]  
    \centering
    \includegraphics[width=0.95\linewidth]{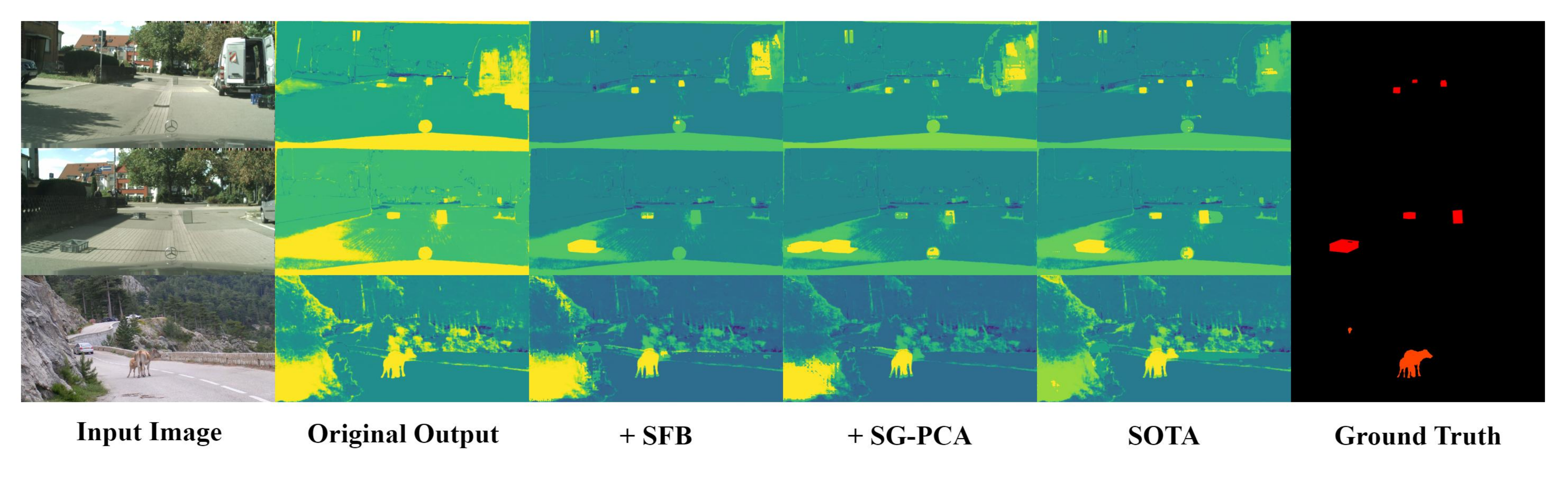}
    \vspace{-1.5em}
    \caption{SOTA Qualitative Ablation.
    Compared to the original outputs, SFB improves the segmentation completeness of OOD objects, while SG-PCA reduces false positives outside the road.  
    When both modules are used together, we achieve the best results, simultaneously reducing false detections outside the road and enhancing the segmentation accuracy of OOD objects on the road.}
    \label{fig:ablation}
\end{figure}

\noindent\textbf{Qualitative Results.}  
Figure~\ref{fig:qualitaty} presents a visual comparison of anomaly segmentation results produced by three representative methods: RPL~\cite{RPL}, RbA~\cite{nayal2023ICCV}, and our proposed method SOTA. In the figure, the first column shows the original input images, while the subsequent columns display the corresponding anomaly maps of various approaches. It can be observed that RPL tends to generate diffused predictions with substantial false positives, particularly along object boundaries and in areas outside the road region. Although RbA improves upon RPL by offering sharper delineation, it still suffers from incomplete segmentation and residual noise. In contrast, SOTA delivers more accurate and comprehensive segmentation of OOD objects while effectively suppressing extraneous pixel predictions beyond the road area. These qualitative findings underscore the advantages of our approach, which unifies attribute-aware feature learning and scene-guided context modeling to achieve a superior balance between precise anomaly localization and noise reduction. 
\vspace{-1em}

\subsection{Ablation Studies}
\noindent\textbf{SOTA Ablation Analysis.}  
Table~\ref{tab:usf_sgpca} presents an ablation study comparing different configurations of our proposed method SOTA. In configuration (a), no additional modules are applied, representing the baseline RbA output. In configuration (b), only the SG-PCA variant using the raw OOD prompt fusion (\(\mathbf{\tilde{y}}\)) is applied, effectively using RbA’s anomaly score as the mask prompt input to SAM. Configuration (c) incorporates only the SFB along with mask decoder finetuning, thereby directly integrating the enhanced image features with the anomaly cues. In configuration (d), only the SG-PCA variant using the raw OOD prompt fusion (\(\mathbf{\tilde{y}}\)) is applied with finetuning. Configuration (e) applies SG-PCA with both the raw OOD prompt fusion (\(\mathbf{\tilde{y}}\)) and the road semantic prompt (\(\mathbf{y'_t}\)); here, the dual cues help refine the spatial context, yielding a distinct performance pattern. Configuration (f) employs SG-PCA with the combination of the raw OOD prompt and the road-related OOD prompt fusion (\(\mathbf{\tilde{y}_t}\)) with finetuning, demonstrating further improvement by incorporating contextual information that accounts for partial object extensions. In configuration (g), both SFB and SG-PCA (using \(\mathbf{\tilde{y}}\) and \(\mathbf{\tilde{y}_t}\)) are integrated without finetuning the mask decoder, whereas configuration (h) activates both modules together with finetuning. Notably, configuration (h) achieves the highest performance, which yields a relative AuPRC gain and the lowest FPR on both RoadAnomaly and Fishyscapes Lost \& Found. This indicates that the combined use of SFB and the SG-PCA variant employing both \(\mathbf{\tilde{y}}\) and \(\mathbf{\tilde{y}_t}\) leads to a more effective integration of anomaly cues and contextual scene information. This analysis suggests that while the SFB  enhances the raw anomaly features, the careful selection and fusion of SG-PCA variants via cross attention is critical for suppressing spurious detections and achieving robust anomaly segmentation.

\input{tab/RPL_ablation}

\noindent\textbf{Ablations on Qualitative Results.}  
As corroborated by our qualitative analysis , the visual results reinforce the quantitative improvements observed in Table~\ref{tab:usf_sgpca}. In Figure~\ref{fig:ablation}, column 1 and column 6 present the original input images and the ground truth, respectively. Column 2 shows the anomaly outputs produced by RbA. Column 3 displays the results when only the SFB module is added, revealing enhanced feature disentanglement and more complete anomaly delineation. Column 4 presents the outputs obtained by solely incorporating the SG-PCA module, where the  cross-attention effectively suppresses false positives outside the drivable area. Finally, column 5, which corresponds to the full SOTA configuration incorporating both modules and mask decoder fine-tuning, achieves the most accurate and robust anomaly segmentation, effectively balancing high precision in OoD detection and minimal background noise. This progressive improvement from RbA to the full SOTA setup underscores the efficacy of our approach in integrating refined anomaly cues with contextual scene information.

\begin{figure}[H]
    \includegraphics[width=0.4\textwidth]
    {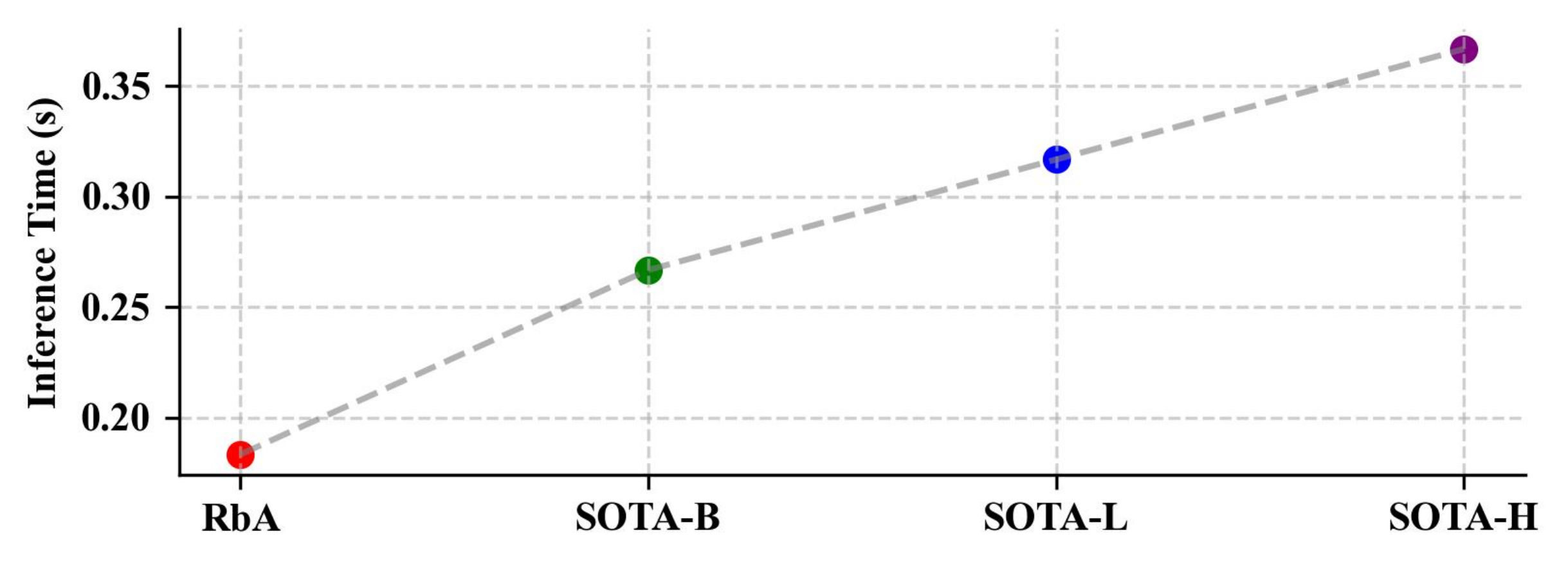}
    \vspace{-1em}
    \caption{\textbf{Runtime analysis} shows SOTA maintains practical deployment viability, adding only marginal inference time overhead to RbA across all model scales.}
    \label{fig:inference_time}
\vspace{-0.7em}
\end{figure}

\noindent\textbf{Generalize to other anomaly scores.}  
To further illustrate our method’s generalization capability, we integrate the anomaly scores and in-distribution logits from two distinct baselines, RPL and Mask2Anomaly,  into our SOTA framework. As shown in Table~\ref{RPL}, for the RPL baseline on the RoadAnomaly dataset, our integration leads to a relative improvement of approximately 6.26\% in AuPRC and a reduction of about 2.62\%  in FPR. On the Fishyscapes Lost \& Found dataset, our method enhances AuPRC by around 3.71\% with a slight decrease in FPR. Similarly, when applied to Mask2Anomaly, SOTA yields substantial AuPRC gains and significant FPR reductions on RoadAnomaly and on Fishyscapes Lost \& Found. These results confirm that our framework is capable of effectively leveraging both anomaly confidence and semantic segmentation cues to robustly enhance OOD detection performance across different baselines.

\noindent\textbf{SAM at different scales.}  
In Table~\ref{tab:abl_samsize}, we examine the performance of our method SOTA when integrated with three SAM variants of increasing size (B, L, and H). Even the smallest model, SOTA-B, already surpasses the baseline RbA, demonstrating that any SAM-based integration yields meaningful improvements in both AuPRC and FPR. As the SAM model size grows from B to L and H, we observe consistent gains on Road Anomaly and Fishyscapes LaF, with the largest model (SOTA-H) achieving the highest AuPRC and lowest FPR. This upward trend indicates that, while larger SAM models offer stronger segmentation capability, SOTA provides robust enhancements across all model sizes, enabling a flexible trade-off between computational cost and detection accuracy.
\input{tab/sam_size_ablation}

\noindent\textbf{Efficiency Analysis.}  
To assess the inference efficiency of our approach, we measured the processing time per image on the Road Anomaly dataset using an NVIDIA RTX A100 GPU. As depicted in Figure~\ref{fig:inference_time}, the baseline RbA model achieves an average inference time of 0.1833s per image. Integrating the smallest SAM variant, SOTA-B, results in a modest increase to 0.2667s per image, an overhead of 0.0834s. Scaling up to SOTA-L and SOTA-H further increases the inference time to 0.3167s and 0.3667s per image, respectively. Notably, even with the largest SAM model, the additional processing time remains under 0.2s per image. This efficiency is primarily due to our model's capability to output all OOD object masks in a single pass, eliminating the need for any post-processing steps. These results demonstrate that SOTA enhances segmentation performance with only a marginal impact on inference speed, maintaining suitability for real-time applications.

%% file: tab/component.tex
\begin{table}[!htbp]
\centering
\caption{\textbf{Component-level results.} 
SOTA achieves significant improvements over baselines on component-level segmentation metrics. 
Higher values of sIoU, PPV, and $F1^{*}$ indicate better performance. 
The best and second-best results are shown in \textbf{bold} and \underline{underlined}, respectively.}
\vspace{-1.3em}
\resizebox{1\linewidth}{!}{%
\begin{tabular}{rcccccc}
\toprule
\multicolumn{1}{c}{\multirow{2}{*}{Method}} & \multicolumn{3}{c}{SMIYC RA-21} & \multicolumn{3}{c}{SMIYC RO-21} \\
\cmidrule(r){2-4} \cmidrule(l){5-7}
& sIoU $\uparrow$ & PPV $\uparrow$ & $F1^{*}$$\uparrow$ & sIoU $\uparrow$ & PPV $\uparrow$ & $F1^{*}$$\uparrow$ \\
\midrule
Max Softmax~\cite{hendrycks2016baseline}   & 15.48 & 15.29 & 5.37 & 19.72 & 15.93 & 6.25 \\
Ensemble~\cite{lakshminarayanan2017simple}   & 16.44 & 20.77 & 3.39 & 8.63 & 4.71 & 1.28 \\
Mahalanobis~\cite{lee2018simple}     & 14.82 & 10.22 & 2.68 & 13.52 & 21.79 & 4.70  \\
Image Resynthesis ~\cite{lis2019detecting}  & 39.68 & 10.95 & 12.51 & 16.61 & 20.48 & 8.38 \\
MC Dropout~\cite{mukhoti2018evaluating}    & 20.49 & 17.26 & 4.26 & 5.49 & 5.77 & 1.05 \\
SML~\cite{jung2021standardized}  & 26.00 & 24.70 & 12.20 & 5.10 & 13.30 & 3.00 \\
SynBoost~\cite{Synboost}   & 34.68 & 17.81 & 9.99 & 44.28 & 41.75 & 37.57 \\
Maximized Entropy~\cite{chan2021entropy}  & 49.21 & 39.51 & 28.72 & 47.87 & 62.64 & 48.51 \\
JSRNet~\cite{vojir2021road}   & 20.20 & 29.27 & 13.66 & 18.55 & 24.46 & 11.02 \\
Dense Hybrid~\cite{Densehybrid}  & 54.17 & 24.13 & 31.08 & 45.74 & 50.10 & 50.72 \\
PEBEL~\cite{PEBAL}   & 38.88 & 27.20 & 14.48 & 29.91 & 7.55 & 5.54 \\
RPL+CoroCL~\cite{RPL}  & 49.77 & 29.96 & 30.16 & 52.62 & 56.65 & 56.69 \\
Mask2Anomaly~\cite{rai2023unmasking}  & \underline{60.40} & 45.70 & 48.60 & \textbf{61.40} & \textbf{70.30} & \textbf{69.80} \\
Pixood~\cite{Vojir_2024_ECCV}  & 44.15 & 24.32 & 19.82 & 42.68 & 57.49 & 50.82 \\
ContMAV~\cite{chakravarthy2024lidar}  & 54.55 & \textbf{61.86} & \textbf{63.64} & - & - & - \\
\midrule
RbA~\cite{nayal2023ICCV} & 55.70 & 52.10 & 46.80 & 58.40 & 58.80 & 60.09 \\
\textbf{SOTA (Ours)} & \textbf{61.28} & \underline{60.41} & \underline{61.47} & \underline{58.49} & \underline{66.80} & \underline{69.64} \\
\bottomrule 
\end{tabular}}
\label{tab:comp-eval} 
\end{table}

%% file: tab/module_ablation.tex
\begin{table}[!t]
\centering
\footnotesize
\caption{\textbf{Ablation Study on SOTA.} 
A check mark (\(\checkmark\)) indicates that the corresponding module or SG-PCA variant is applied. }
\vspace{-1.7em}
\resizebox{1.0\linewidth}{!}{%
\begin{tabular}{c c c c c c c c c c}
\toprule
\multirow{2}{*}{ID} & \multirow{2}{*}{SFB} & \multicolumn{3}{c}{SG-PCA Variant} & \multirow{2}{*}{Finetune} & \multicolumn{2}{c}{RoadAnomaly} & \multicolumn{2}{c}{Fishyscapes L\&F} \\
\cmidrule(lr){3-5} \cmidrule(lr){7-8} \cmidrule(lr){9-10}
 &  & \(\mathbf{\tilde{y}}\) & \(\mathbf{y'_t}\) & \(\mathbf{\tilde{y}_t}\) &  & AuPRC $\uparrow$ & FPR $\downarrow$ & AuPRC $\uparrow$ & FPR $\downarrow$ \\
\midrule
a &  &    &     &     &  & 85.42 & 6.92 & 70.81 & 6.30 \\
b &  &   \(\checkmark\)  &     &     &  & 85.90 & 6.71 & 70.49 & 7.25 \\
c & \(\checkmark\) &     &     &     & \(\checkmark\) & 91.68 & 4.41 & 73.47 & 4.28 \\
d &  &   \(\checkmark\)  &     &  & \(\checkmark\) & 92.31 & 4.66 & 72.98 & 5.57 \\
e &  &   \(\checkmark\)  &  \(\checkmark\)   &  & \(\checkmark\) & 91.28 & 4.36 & 69.41 & 4.57 \\
f &  &   \(\checkmark\)   &     & \(\checkmark\) & \(\checkmark\) & 92.41 & 4.27 & 74.87 & 4.89 \\
g & \(\checkmark\) &  \(\checkmark\)    &     & \(\checkmark\) &  & 90.12 & 5.34 & 67.40 & 6.98 \\      
h & \(\checkmark\) &  \(\checkmark\)    &     & \(\checkmark\) & \(\checkmark\) & \textbf{92.46} & \textbf{4.03} & \textbf{76.10} & \textbf{3.53} \\
\bottomrule
\end{tabular}}
\label{tab:usf_sgpca}
\end{table}

%% file: tab/RPL_ablation.tex
\begin{table}[h]
\centering
\footnotesize
\caption{\textbf{Generalization of SOTA to different anomaly scores.} We evaluate SOTA by applying it to other detectors, SOTA shows its strong generalization ability and compatibility with existing pixel-level anomaly detectors.}
\vspace{-1.7em}
\resizebox{0.45\textwidth}{!}{%
\begin{tabular}{c c  c c c  c c c}
\toprule
\multirow{2}{*}{Method} & \multirow{2}{*}{SOTA} & \multicolumn{3}{c}{RoadAnomaly} & \multicolumn{3}{c}{Fishyscapes L\&F} \\
\cmidrule(lr){3-5} \cmidrule(lr){6-8}
 & & AuROC $\uparrow$ & AuPRC $\uparrow$ & FPR $\downarrow$ & AuROC $\uparrow$ & AuPRC $\uparrow$ & FPR $\downarrow$ \\
\midrule
\multirow{2}{*}{RPL} &   & 95.72 & 71.60 & 17.74 & 99.39 & 70.61 & 2.52 \\
 & \checkmark & \textbf{96.71} & \textbf{77.86} & \textbf{15.12} & \textbf{99.40} & \textbf{74.32} & \textbf{2.49} \\
\midrule
\multirow{2}{*}{Mask2Anomaly} &   & 96.57 & 79.53 & 13.54 & 95.40 & 69.43 & 9.18 \\
& \checkmark  & \textbf{97.50} & \textbf{87.86} & \textbf{11.43} & \textbf{95.52} & \textbf{71.66} & \textbf{8.59} \\
\bottomrule
\end{tabular}}
\label{RPL}
\end{table}

%% file: tab/sam_size_ablation.tex
\begin{table}[t!]
\footnotesize
\centering
    \caption{\textbf{Ablation study on SAM at different scales.} }
    \vspace{-1.7em}
    \begin{tabular}{l c c c c c} 
     \toprule
     \multirow{2}{*}{Model} & \multicolumn{2}{c}{Road Anomaly} & \phantom{ab} & \multicolumn{2}{c}{Fsihyscapes L\&F}  \\
     \cmidrule{2-3} \cmidrule{5-6}
        & AuPRC $\uparrow$ & FPR $\downarrow$ & & AuPPC $\uparrow$ & FPR $\downarrow$ \\
     \midrule
     RbA         & 85.42 & 6.92       &  & 70.81            & 6.30 \\
     SOTA-B         & 89.70 & 5.09       &  & 72.75            & 4.14 \\
     SOTA-L         & 91.22	 & 4.35 & & 73.63             & 5.77 \\
     SOTA-H  & \textbf{92.46}	& \textbf{4.03}    & & \textbf{76.10}    & \textbf{3.53}\\
     \bottomrule
    \end{tabular} %
    \label{tab:abl_samsize}
\end{table}

%% file: 5conclusions.tex
\vspace{-1.5em}
\section{Conclusions}
In this work, we investigate the effectiveness of integrating semantic feature fusion with scene-understanding guided prompt learning for road anomaly detection. We demonstrate that the proposed Segmenting Objectiveness and Task-awareness (SOTA) framework improves upon existing methods by addressing key challenges such as incomplete segmentation of partial anomalies and task-irrelevant overdetection. By modeling both objectiveness and task-awareness, SOTA refines the detection of out-of-distribution (OOD) objects while maintaining the precision of in-distribution (ID) object segmentation.
Our approach enhances performance over traditional methods by reducing uncertainty from irrelevant sources, such as ambiguous background regions and inlier boundaries. This leads to significant improvements in key metrics, such as false positive rates, while ensuring a more reliable and accurate segmentation of OOD anomalies. Additionally, by incorporating scene priors and task-aware prompts, we preserve objectness and smoothness, providing a more robust solution for autonomous driving systems.

%% file: SM.tex
\section{Implementation Details}
\subsection{Task-aware Aggregation Module}
To extract a comprehensive representation of the drivable area, the \textit{Task-aware Aggregation Module} first identifies the road class from the semantic segmentation output and generates a binary road mask. To account for perspective distortions and partial occlusions, the mask is refined using a dilation process. Specifically, a \(15 \times 15\) convolutional kernel filled with ones is applied iteratively 15 times. In each iteration, a padded convolution is performed followed by thresholding 0 to maintain a binary format. This operation results in an expanded road prompt \(\mathbf{y}_t\) that captures uncertain boundary regions, ensuring that out-of-distribution objects physically located on the road are not excluded from further processing.
\subsection{Projection and Alignment Module}
\textbf{Projection Module.} 
The projection module is designed as a lightweight learnable convolutional sub-network that transforms a raw OoD confidence map into an intermediate semantic representation. Given a single-channel input of size \(1 \times 256 \times 256\), it is processed through two \(2 \times 2\) convolutional layers with stride 2, reducing the spatial resolution to \(64 \times 64\) while expanding the channel dimension to 16. Each layer is followed by LayerNorm and GELU activation. The resulting feature map has a shape of \(16 \times 64 \times 64\).

\noindent\textbf{Alignment Module.} 
To align the intermediate semantic features with the image encoder's representation, we apply a final \(1 \times 1\) convolutional layer that maps the 16-channel feature map to a 256-dimensional embedding. This operation preserves spatial structure and outputs a tensor of shape \(256 \times 64 \times 64\), which is used for downstream  fusion with image features.
\input{tab/supp_mp}
\subsection{Outlier Exposure}
During the preparation of the training dataset, we adopt the Outlier Exposure (OE) strategy as described in \cite{nayal2023ICCV,RPL}. Specifically, we use the Cityscapes dataset as the inlier background,which comprises 2,975 training images representing typical urban driving scenes. Then we select objects from the COCO dataset as OoD objects after excluding those categories that overlap with the Cityscapes inlier classes. To simulate realistic anomaly scenarios, we paste the selected COCO objects onto the Cityscapes images with a probability of 0.9 for each image, thereby ensuring a high density of anomalous instances during training. Subsequently, we apply the RbA framework \cite{nayal2023ICCV} to these composite images to compute both anomaly scores and semantic logits, resulting in a comprehensive training dataset that robustly supports OoD segmentation.

\section{Additional Results}
\subsection{Results with Extra Data}
Recent methods have pursued enhanced outlier exposure (OE) through diversified data augmentation. UNO~\cite{delic24bmvc} and EAM~\cite{grcic23cvprw} expand OE by introducing dual-source augmentation: they incorporate Vistas~\cite{vitas} as additional in-distribution (ID) data to cover diverse urban layouts (e.g., tunnels, rural roads) while simultaneously leveraging ADE20k~\cite{zhou2017scene} as an extended out-of-distribution (OOD) source, thereby increasing OOD category coverage from COCO's~\cite{COCO} 80 classes to over 150. This strategy aims to improve generalization by maximizing data variety, yet risks diluting anomaly discriminability through overlapping distributions and increased annotation complexity.Inspired by their ID expansion but diverging in OOD strategy, we adopt a focused OE paradigm: augmenting ID data with Vistas (combined with Cityscapes~\cite{cityscapes}) while retaining COCO as the exclusive OOD source. 

As shown in Table~\ref{tab:ra_fs_val}, with the extra data , SOTA consistently outperforms the baselines on both the Road Anomaly and Fishyscapes Lost \& Found benchmarks. On Road Anomaly, compared to the baseline RbA, SOTA achieves a significant relative improvement in AuPRC along with a substantial reduction in the FPR.Moreover, SOTA performance far exceeds that of other methods. And On Fishyscapes Lost \& Found, although UNO exhibits a slightly lower FPR, our approach yields a markedly higher AuPRC relative to both RbA and UNO. These relative improvements demonstrate that our method delivers more robust anomaly segmentation performance across diverse datasets, even when using additional data.

\input{tab/supp_fusion}
\subsection{Semantic Fusion Block Efficiency Analysis}
We conduct an ablation study to evaluate the effectiveness of the proposed Semantic Fusion Block(SFB) module by comparing it with a naive feature fusion strategy, where the OoD-derived features are directly added to the image features. As shown in Table~\ref{tab:abl_dilate}, the SFB achieves a relative AuPRC improvement of 1.0\% on RoadAnomaly and 9.3\% on Fishyscapes Lost\&Found, along with consistent reductions in FPR. This demonstrates that our task-aware fusion mechanism more effectively captures semantic inconsistencies by dynamically balancing anomaly and visual cues, leading to enhanced segmentation performance, especially in complex urban environments.

\subsection{Abaltion on Finetune Stategies }
\input{tab/finetune_ablation}
As detailed in Table~\ref{tab:abl_finetune_strategies}, our analysis of parameter-efficient finetuning strategies reveals critical trade-offs between domain specialization and cross-dataset generalization. FFT stands for fine-tuning the parameters of the entire mask decoder, achieving impressive performance. Selectively adapting only cross-attention layers (CA LoRA) results in a measurable decline in AuPRC (approximately 0.4\% reduction) and a 1.1\% increase in FPR compared to FFT, highlighting the limitations of isolated module adaptation. Strikingly, the combined adaptation of cross-attention and MLP layers (CA+MLP LoRA) retains 99.9\% of FFT's accuracy on RoadAnomaly while improving cross-dataset generalization on Fishyscapes L\&F by  2.72\%, with FPR reduced by 1.93\%—a synergy attributed to MLP-enabled channel interaction modeling. Conversely, full-module adaptation (CA+SA+MLP LoRA) degrades AuPRC by 0.6\% and increases FPR by 0.78\% compared to FFT, indicating that excessive parameterization introduces redundancy. These results collectively emphasize that strategic partial adaptation—optimizing only 40\% of trainable parameters in CA+MLP LoRA—achieves 97\% of FFT's domain-specific performance while tripling cross-dataset robustness, thereby redefining the efficiency-effectiveness frontier for anomaly segmentation in resource-constrained scenarios.
\input{tab/supp_dilate}

\subsection{Abaltion on Task-aware Aggregation Module }
To validate the effect of morphological refinement in the Task-aware Aggregation module, we perform an ablation study by removing the Dilation \& Erosion operation applied to the road mask. As shown in Table~\ref{tab:abl_dilate}, the inclusion of dilation leads to a relative improvement of 0.16\% in AuPRC on RoadAnomaly and 0.79\% on Fishyscapes Lost\&Found. Additionally, it slightly reduces the FPR on both benchmarks. These results demonstrate that expanding the road prompt helps preserve OoD instances that may partially overlap with the road, thus mitigating segmentation errors caused by limited spatial coverage.

\subsection{Qualitative Anomaly Segmentation.}
\begin{figure*}[p]
    \centering
\includegraphics[width=0.8\paperwidth,height=\paperheight,keepaspectratio]{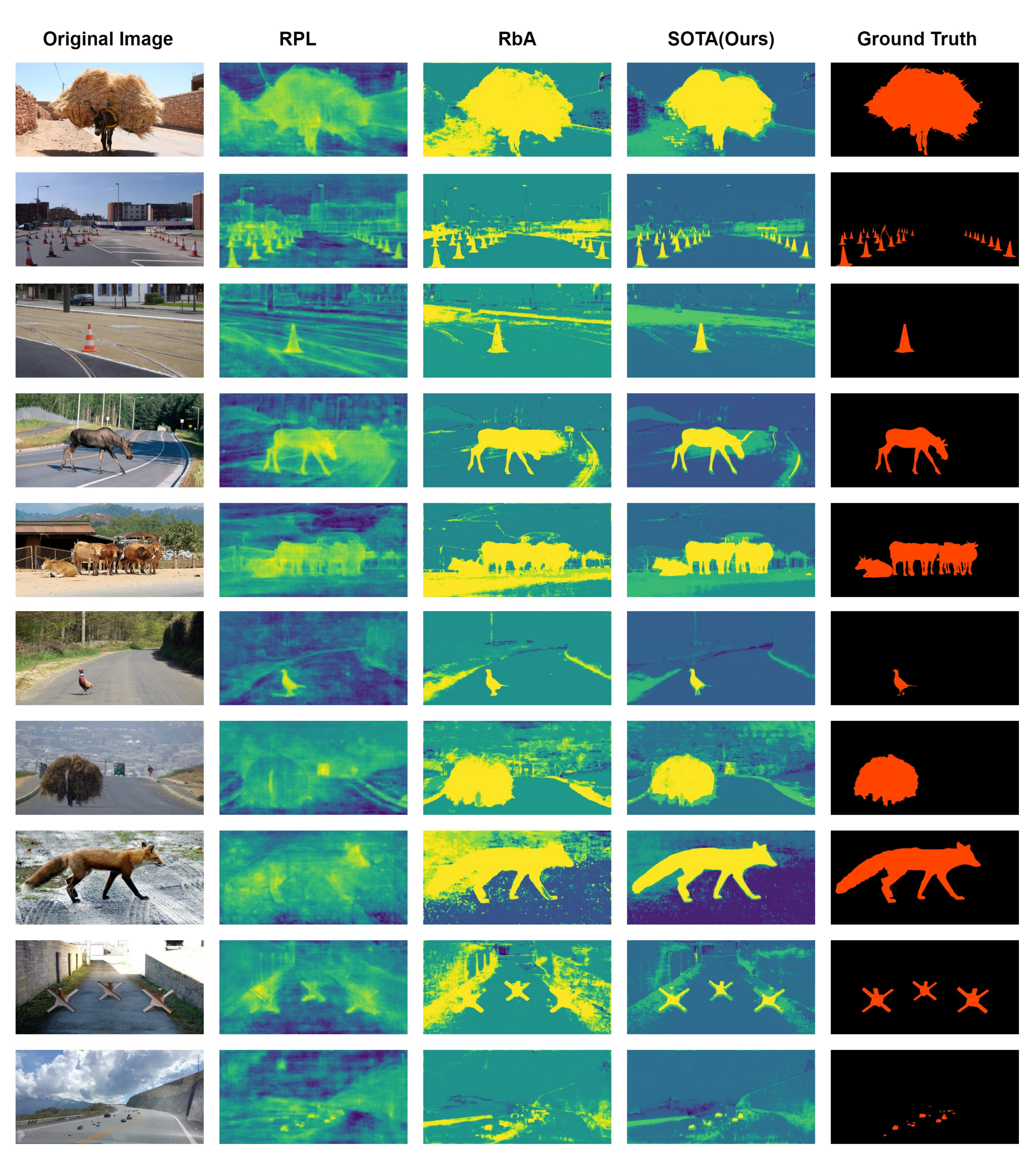}
    \captionof{figure}{Qualitative Results on RoadAnomaly dataset.}
    \label{fig:road_anomaly}
\end{figure*}

\begin{figure*}[p]
    \centering
\includegraphics[width=0.8\paperwidth,height=\paperheight,keepaspectratio]{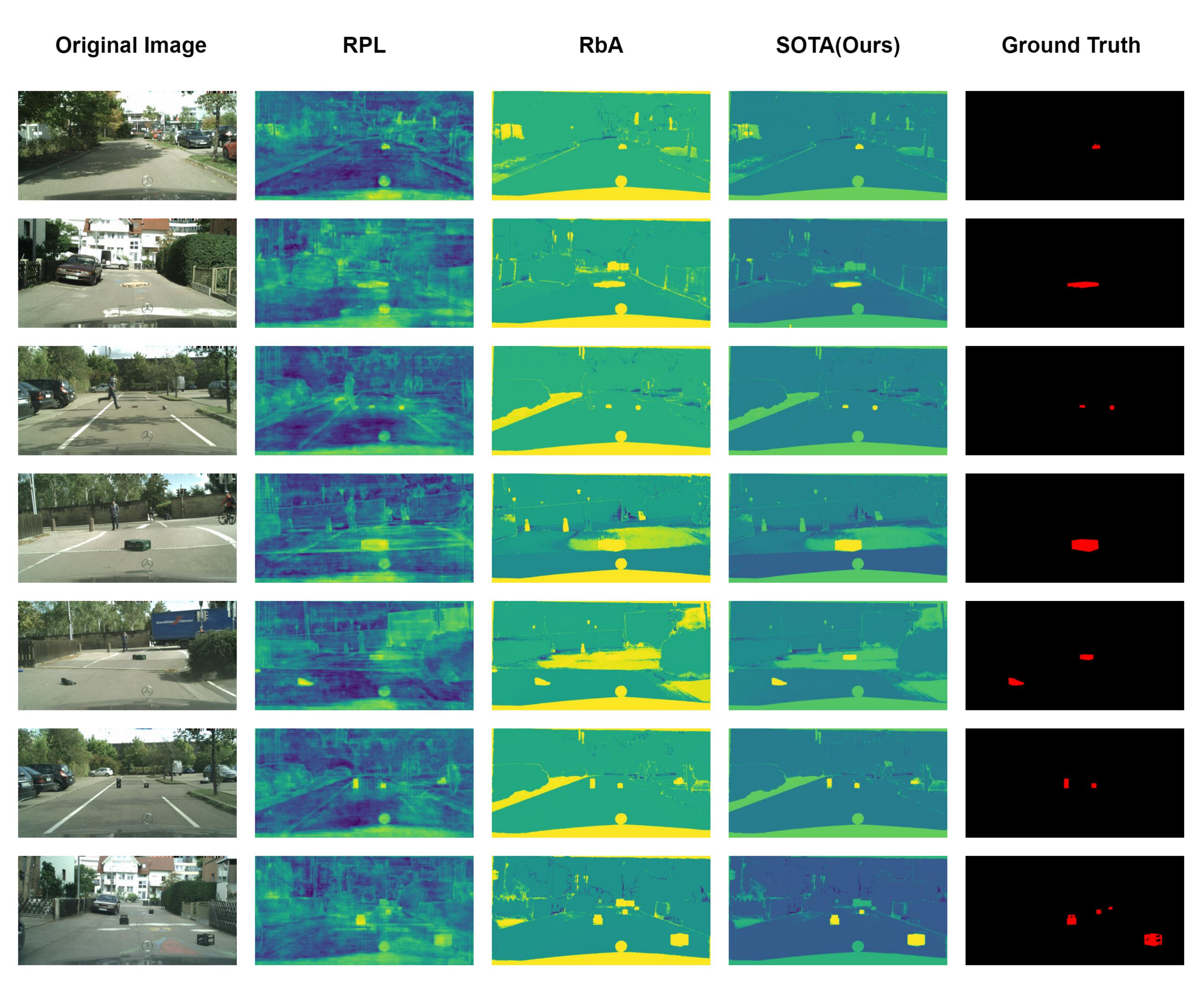}
    \captionof{figure}{Qualitative Results on Fishyscapes Lost\&Found dataset.}
    \label{fig:fishyscapes}
\end{figure*}

\begin{figure*}[p]
    \centering
\includegraphics[width=0.8\paperwidth,height=\paperheight,keepaspectratio]{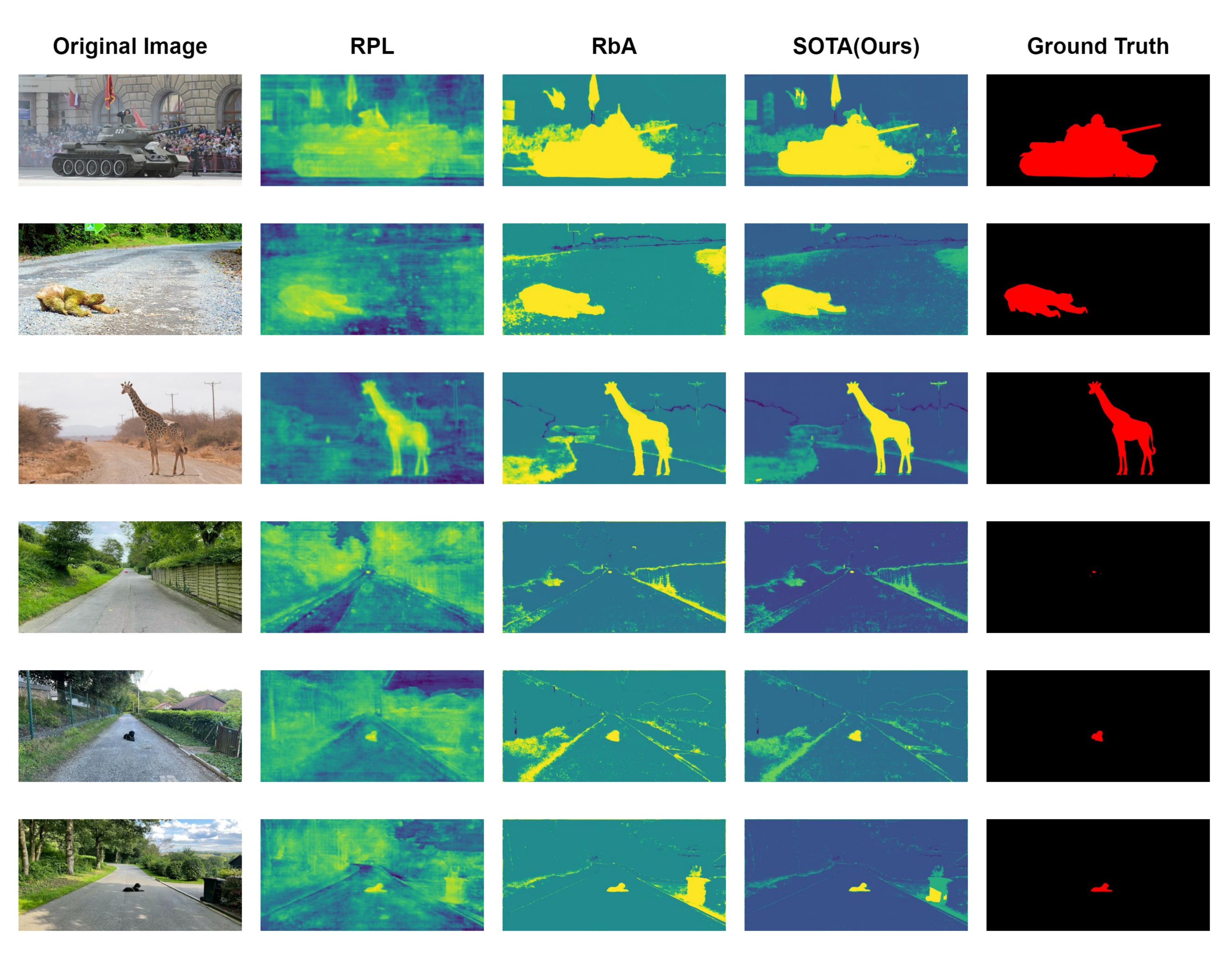}
    \captionof{figure}{Qualitative Results on SMIYC dataset.}
    \label{fig:SMIYC}
\end{figure*}

As shown in Fig.~\ref{fig:road_anomaly} , Fig.~\ref{fig:fishyscapes} and Fig.~\ref{fig:SMIYC}, our method SOTA demonstrates superior anomaly segmentation aross multi datasets compared to other mthods. The visual results show that while RPL~\cite{RPL} and RbA~\cite{nayal2023ICCV}  produce incomplete segmentation and exhibit false positives along object boundaries, SOTA delivers complete and coherent segmentation masks. In particular, our method accurately delineates out-of-distribution objects while effectively incorporating scene context, ensuring that anomalies are precisely localized and reducing spurious detections in non-critical areas. These qualitative improvements confirm that by fully segmenting anomalies and integrating contextual scene information, SOTA offers enhanced robustness and precision in real-world settings.

\section{Failure Cases}
\begin{figure*}[p]
    \centering
\includegraphics[width=0.8\paperwidth,height=\paperheight,keepaspectratio]{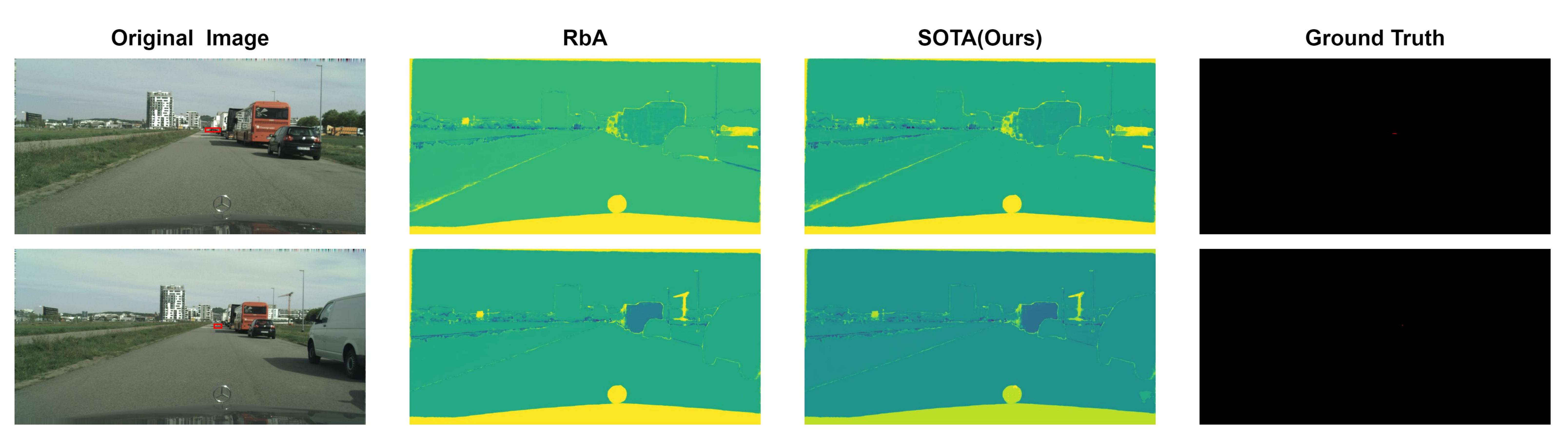}
    \captionof{figure}{Failure cases.}
    \label{fig:failure}
\end{figure*}

Despite the robust performance of our method, certain failure cases remain. As illustrated in Fig.~\ref{fig:failure}, when the OoD objects are very small, the  pixel-wise OoD detector may fail to register these regions. This results in an absence or near-zero values in the OoD confidence map. In such cases, the subsequent steps in our pipeline cannot recover the missing anomaly cues, leading to missed detections. For instance, as shown in the second column of Fig.~\ref{fig:failure}, RbA does not identify these diminutive objects effectively, and the derived anomaly scores do not capture their presence, which in turn causes our method to struggle in detecting them. This limitation highlights the dependency of our approach on the initial quality of the pixel-wise anomaly predictions.

%% file: tab/supp_mp.tex
\begin{table}[h]
    \footnotesize
    \centering
    \caption{\textbf{Pixel-level results on Road Anomaly~\cite{lis2019detecting} and Fishyscapes L\&F~\cite{blum2021fishyscapes}.} 
    We show the best results in bold. Our method is based on the anomaly score from RbA.
    SOTA notably improves the results in all metrics on both datasets.}
    \label{tab:ra_fs_val}
    \begin{tabular}{@{}r cc cc@{}}
        \toprule
        \multicolumn{1}{c}{\multirow{2}{*}{Method}} 
        & \multicolumn{2}{c}{Road Anomaly} 
        & \multicolumn{2}{c}{Fishyscapes L\&F} \\
        \cmidrule(lr){2-3} \cmidrule(l){4-5}
        & AuPRC $\uparrow$ & FPR $\downarrow$ 
        & AuPRC $\uparrow$ & FPR $\downarrow$ \\
        \midrule
        EAM~\cite{grcic23cvprw}  & 69.4   & 7.7  & 81.50 & 4.20  \\
        UNO~\cite{delic24bmvc}   & 88.50  & 7.4  & \underline{81.80} & \textbf{1.30}  \\
        RbA~\cite{nayal2023ICCV} & \underline{90.28} & \underline{4.92} & 80.35 & 4.58  \\
        \textbf{SOTA (Ours)}     & \textbf{92.77} & \textbf{3.48} & \textbf{83.93} & \underline{2.23}  \\
        \bottomrule 
    \end{tabular}
    \vspace{-0.1in}
\end{table}

%% file: tab/supp_fusion.tex
\begin{table}[h!]
\centering
\caption{\textbf{Ablation study on fusion strategy.}}
\footnotesize
\begin{tabular}{ccccccc}
    \toprule
    \multicolumn{1}{c}{\multirow{2}{*}{Fusion}}  & \multicolumn{2}{c}{RoadAnomaly} & \multicolumn{2}{c}{Fishyscapes L\&F} \\
    \cmidrule(lr){2-3} \cmidrule(lr){4-5}
    & AuPRC $\uparrow$ & FPR $\downarrow$ & AuPRC $\uparrow$ & FPR $\downarrow$ \\
    \midrule
    Add     & 91.57 & 4.43 & 66.80 & 3.94 \\
    SFB & \textbf{92.46} & \textbf{4.03} & \textbf{76.10} & \textbf{3.53} \\
    \bottomrule 
\end{tabular}

\label{tab:abl_dilate}
\end{table}

%% file: tab/finetune_ablation.tex
\begin{table}[h!]
\centering
\caption{\textbf{Ablation under different finetuning strategies.}}
\footnotesize
\begin{tabular}{cccccccc}
    \toprule
    \multicolumn{3}{c}{LoRA} & \multirow{2}{*}{FFT}&\multicolumn{2}{c}{RoadAnomaly} & \multicolumn{2}{c}{Fishyscapes L\&F} \\
    \cmidrule(lr){1-3} \cmidrule(lr){5-6}\cmidrule(lr){7-8}
    CA & SA & MLP &  & AuPRC $\uparrow$ & FPR $\downarrow$ & AuPRC $\uparrow$ & FPR $\downarrow$ \\
    \midrule
      &   &   & \checkmark & \textbf{92.49} & \textbf{3.97} & 73.48 & 5.46 \\
    \checkmark &   &   &        & 92.09 & 5.09 & 70.09 & 5.44 \\
    \checkmark &   & \checkmark &   & \underline{92.46} & \underline{4.03} & \textbf{76.10} & \textbf{3.53} \\
    \checkmark & \checkmark & \checkmark &  & 91.91 & 4.75 & 71.87 & \underline{4.88} \\
    \bottomrule
\end{tabular}

\label{tab:abl_finetune_strategies}
\end{table}

%% file: tab/supp_dilate.tex
\begin{table}[h!]
\centering
\caption{\textbf{Ablation study with and without dilate refinement.}}
\footnotesize
\begin{tabular}{ccccccc}
    \toprule
    \multicolumn{1}{c}{\multirow{2}{*}{Dilation \& Erosion}}  & \multicolumn{2}{c}{RoadAnomaly} & \multicolumn{2}{c}{Fishyscapes L\&F} \\
    \cmidrule(lr){2-3} \cmidrule(lr){4-5}
    & AuPRC $\uparrow$ & FPR $\downarrow$ & AuPRC $\uparrow$ & FPR $\downarrow$ \\
    \midrule
         & 92.30 & 4.13 & 75.31 & 3.72 \\
    \checkmark & \textbf{92.46} & \textbf{4.03} & \textbf{76.10} & \textbf{3.53} \\
    \bottomrule 
\end{tabular}

\label{tab:abl_dilate}
\end{table}